\documentclass[10pt,twocolumn,letterpaper]{article}
\usepackage{wacv}
\usepackage{times}
\usepackage{epsfig}
\usepackage{graphicx}
\usepackage{amsmath}
\usepackage{amssymb}
\usepackage{booktabs}

\def\wacvPaperID{790} 

\wacvapplicationstrack 

\wacvfinalcopy 

\ifwacvfinal
\usepackage[breaklinks=true,bookmarks=false]{hyperref}
\else
\usepackage[pagebackref=true,breaklinks=true,colorlinks,bookmarks=false]{hyperref}
\fi

\begin{document}

\title{Objects as Spatio-Temporal 2.5D points}

\author{Paridhi Singh\\
{\tt\small paridhi@ridecell}
\and
Gaurav Singh\\
{\tt\small gaurav@ridecell.com}
\and
Arun Kumar\\
{\tt\small arun.kumar@ridecell.com}
}

\maketitle

\vspace{-0.3in}
\begin{abstract}
 \vspace{-0.1in}
 Determining accurate bird's eye view (BEV) positions of objects and tracks in a scene is vital for various perception tasks including object interactions mapping, scenario extraction etc., however, the level of supervision required to accomplish that is extremely challenging to procure. We propose a light-weight, weakly supervised method to estimate 3D position of objects by jointly learning to regress the 2D object detections and scene's depth prediction in a single feed-forward pass of a network. Our proposed method extends a center-point based single-shot object detector \cite{zhou2019objects}, and introduces a novel object representation where each object is modeled as a BEV point spatio-temporally, without the need of any 3D or BEV annotations for training and LiDAR data at query time. The approach leverages readily available 2D object supervision along with LiDAR point clouds (used only during training) to jointly train a single network, that learns to predict 2D object detection alongside the whole scene's depth, to spatio-temporally model object tracks as points in BEV. The proposed method is computationally over $\sim$10x efficient compared to recent SOTA approaches \cite{bhat2021adabins, wang2019pseudo} while achieving comparable accuracies on KITTI tracking benchmark.
\end{abstract}
\vspace{-0.2in}

\section{Introduction}
\vspace{-0.05in}
\begin{figure}[hbt!]

 \includegraphics[width=0.5\textwidth]{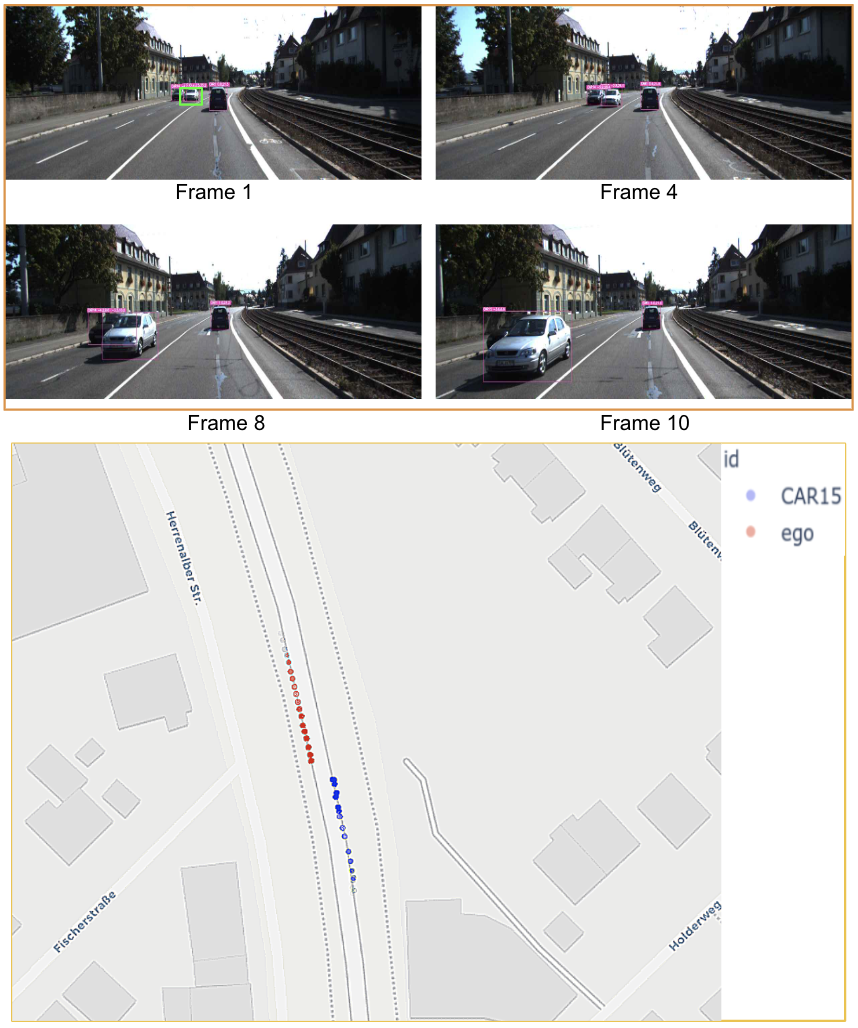}
 \vspace{-2 mm}
 \caption{Qualitative representation of extracted interactions between ego vehicle and an actor of interest (marked in green in Frame 1): Sequence of images with its 2D object detections (top), and object 3D positions plotted on OpenStreetMaps~\cite{osm2004} (bottom). The red colored object track represents the positions of the ego vehicle, where the other actor's track (estimated relative to ego vehicle's GPS position using the proposed method) is plotted in blue.}
 \vspace{-0.3in}
\end{figure}

While the most common use of computer vision for autonomous vehicles (AV) is for the real-time perception and navigation systems of the vehicle, there are various other applications that utilize the collected data by querying large amounts of data, such as data curation, automated data tagging {\it etc.}, performing analytics on the drive data for use cases like creating HD maps, compliance adherence for regulatory bodies, risk analysis for insurance etc. 
These ADAS \& AV fleets collect enormous amounts of data. Thus, unlike most other computer vision problems where the primary challenge is not having enough data, in this case the problem is having too much data and lacking sophisticated solutions to effectively utilize them.  
Traditionally these applications are run on the server, but more recently there was a rising need for in-vehicle data processing to prune away unnecessary data instead of sending them to the cloud which incurs significant network and cloud costs to transfer, store and process them.
Most of the data analysis for such applications revolves around identifying objects in 3D and modeling their 3D interactions over time. There are two major challenges in performing such large scale data analysis (1) the network needs to be light weight, capable of running on low compute and resource-constraint edge environments and (2) it has to be scale-able and extendable across a large number of object classes with minimal need for annotations. 

The paper proposes a computer vision method that is (1) light weight and can run on any hardware environment from 2GB nano board, snapdragon chips to dedicated GPU Servers without considerable differences in performance, (2) can learn to predict \& track 3D positions of the objects without requiring any 3D annotations during training or expensive sensors during test time. 

While determining the full 3D bounds along with orientation of the objects are critical for tasks like autonomous navigation, they are certainly an overkill for an application like automated label generation which is running on an edge device. For instance, querying interesting driving scenarios or interactions such as cut-ins, lane changes, pedestrian jaywalking etc. do not require 3D size of road objects or their orientations, they only require coarse 3D positions of the objects reliably tracked over time. The paper proposes a simple and effective method that is scale-able for performing large scale data analysis with minimal need for expensive sensors or 3D annotations, that can also scale across different hardware environments.


Instead of using expensive 3D networks, we propose to equip 2D object detection techniques with ability to reason the 3D scene geometry and object's motion dynamics. We introduce in this paper a novel architecture for jointly learning to detect, track and model objects in BEV, using only 2D bounding box annotations and tracks in a weakly supervised manner. This proposed method is aimed squarely toward edge and offboard perception tasks such as generating automated tags for objects, or model object-object interactions to identify abnormal/salient driving pattern etc. from camera only data like data from dashboard cameras. An example of this use case and approach can be seen in Fig.1, where target vehicle tracks are represented in BEV space along with ego vehicle tracks and the representation is used to extract important interaction scenarios. Fig. 1 shows tracks generated using the method described in this paper. Note that notation of objects as 2.5D signify that points lie on the surface of the object, thus are not necessarily the 3D center of objects. But, in context to representing objects or modeling object interactions, they are essentially the same

Our method leverages LiDAR data for training monocular network for such offline perception tasks. The sparse depth information available via LiDAR can be utilized to train perception networks to model objects and interactions among them in BEV efficiently, while using only camera videos at test time. The goal of this work is to demonstrate that the method can scale for large scale 3D data analysis with little need for annotation. The use of LiDAR for training depth prediction is also optional, where self-supervised depth prediction techniques with scale imparted from the Vehicle's IMU data has been shown to work well~\cite{guizilini20203d}. But, we limit the scope of this paper to only demonstrate 3D object interactions using LiDAR weak-supervision only during training, whereas camera suffices during test time.

We extend a 2D center-point based object detection technique~\cite{zhou2019objects} to learn the whole scene's depth. By doing so, we force the network to learn global scene-level scale to reason the scene's 3D structure more accurately. Traditionally, object detection and depth prediction tasks are often employed in isolation, and often thought of as different problems. While a few attempts have been made to combine them~\cite{you2019pseudo}, the methods often stack one method after the other. We combine both tasks to be learnt by a single network, in order to enhance the object detection accuracy by instilling the local and global scene geometry. In addition, the proposed method forces the network to predict the scale of objects more accurately, as the model looks at the whole scene to learn to predict the depth of an object, instead of looking at each object in isolation~\cite{liu2020smoke}. 
\section{Related Work}
\vspace{-0.05in}
Existing literature in 2D+ object detection can be categorized into LiDAR-based methods~\cite{ku2018joint, simon2018complex, lang2019pointpillars},  that take raw or processed LiDAR 3D points as inputs, monocular (camera-based) methods~\cite{liu2020smoke, zhou2019objects}, or a hybrid of the two~\cite{wang2019pseudo, you2019pseudo, weng2019monocular}. 
While LiDAR based methods are generally more accurate in determining the 3D (x, y \& z) and BEV positions of objects in a scene, they also require that data from LiDAR sensor during inference, which is quite expensive and often unavailable for our offline perception use cases. On the other hand, cameras are relatively cheaper sensors, but they lose 3D information due to the way an image is captured while preserving the rich textural cues.
Thus, evidently, monocular camera-based methods fall short in predicting the accurate depth (z) of objects in a scene in comparison with their LiDAR-based counterparts, but are relatively more accurate when it comes to predicting (x \& y) of the scene. Attempts have been made to combine both modalities, in a hybrid setting, to exploit the best of both worlds~\cite{wang2019pseudo}. But most of them require the LiDAR data as input at query time. While there are a few methods~\cite{wang2019pseudo, you2019pseudo, weng2019monocular} that predict depth and use them for LiDAR object detection, they rely heavily on the availability of 3D object annotations. 3D object annotations are challenging for two main reasons, they are challenging to annotate, and they often cover very less object classes.

\begin{figure*}[hbt!]
 \centering
 \includegraphics[width=1.0\textwidth, keepaspectratio]{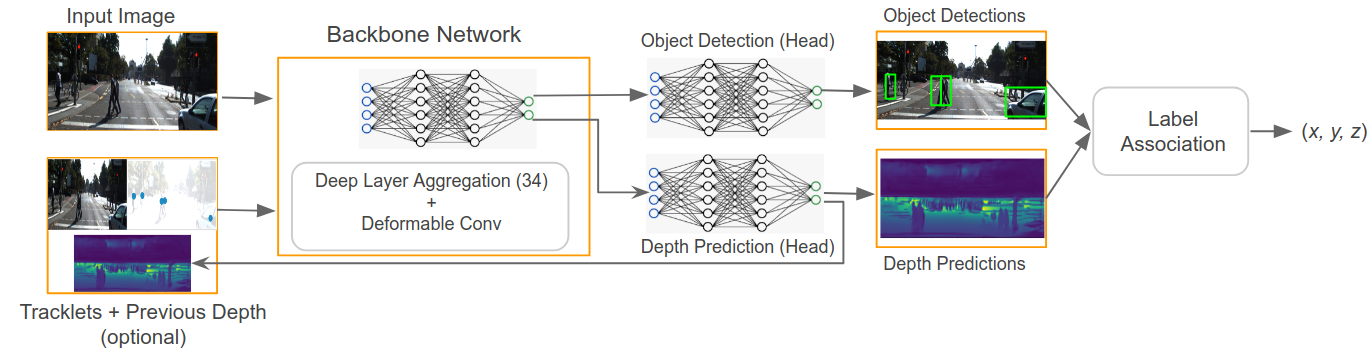}
 \caption{Pipeline of the proposed framework. Given input image(s) and LiDAR point clouds, a Deep Layer Aggregation-based backbone with deformable convolutional layers, encodes the input image, followed by two heads for depth prediction and for object detection respectively. Use of tracklets and the previous frame depth output is an additional feature, the proposed method has an inbuilt tracking pipeline (of ~\cite{zhou2020tracking} in order to smoothen the object detections.}
 \vspace{-0.2in}
\end{figure*}

The method proposed follows a hybrid approach, uses both camera data and LiDAR point cloud data for training, but there are two major differences (1) the LiDAR data is not used as an input at test time and thus LiDAR sensor is not required in the test vehicle, and (2) the method relies on camera 2D object detections alone instead of 3D annotations (as in~\cite{wang2019pseudo, you2019pseudo}). 3D annotations are expensive to annotate and are often restricted to a small number of classes, and (3) the method combines object detection and tracking into a single step feedforward pass ensuring spatio-temporally smooth object detections. The proposed method is also relevant to~\cite{wang2019pseudo, you2019pseudo}, where instead of building a second object detection network on top of the regression depth, which further adds to the already computationally intensive depth prediction process, we propose a single network that, given a camera image as input, can learn both object detection and dense depth in a multimodal regression setting, where the ground truth LiDAR data is used at training time only to compute depth regression loss. Since, the network uses as input, the camera image alone, at test time, the network can predict both object detections and dense depth of the scene. 



\noindent{\textbf{Monocular 2D Object Detection:}}
 2D Object detection networks can be divided into anchor-based~\cite{ren2015faster, qin2019thundernet, he2019bounding, zhang2018single} or anchor free architectures~\cite{redmon2016you, redmon2018yolov3, law2018cornernet, yin2020center}. Anchor-based architectures are further categorized into either two-staged ~\cite{ren2015faster, qin2019thundernet, he2019bounding, tan2019learning, pang2019libra} or computationally efficient single stage ~\cite{liu2016ssd, liu2018receptive, chen2019towards, zhang2018single}. On the other hand, anchor-free detectors~\cite{redmon2018yolov3, yin2020center} directly find objects either using a key-point or center-point based approaches. 

CornetNet~\cite{law2018cornernet} models each object as a pair of keypoints placed at the corners of the bounding boxes and CenterNet~\cite{yin2020center} combines ~\cite{law2018cornernet} with center-point~\cite{redmon2018yolov3} based approaches by adding an  additional keypoint at the center of the bounding box effectively modeling each object as a keypoint triplet. The proposed method builds over center point-based single-stage architecture~\cite{yin2020center} for 2D object detection.

\noindent{\textbf{Monocular 3D Object Detection:}}
Monocular 3D object detection methods~\cite{chen2016monocular, liu2020smoke, zhou2019objects} regress complete 3D object information such as the center of the object in 3D, offsets of the 3D bounding box along with its orientation, all from a monocular image. 
CenterTrack \cite{zhou2020tracking} builds over CenterNet \cite{zhou2019objects} which models objects as centers, by jointly detecting and tracking object instances in 3D using a tracking conditioned on detection method. The proposed method further extends center-based method~\cite{yin2020center} to spatio-temporally model objects as points in 3D, in a weakly-supervised manner, aiming to achieve 3D representation for objects, without relying on a comprehensive 3D object detection architectures that require laborious and exorbitant bounding boxes and orientation annotations. 

\noindent{\textbf{Monocular Depth Prediction:}}
Most literature for predicting pixel-wise depth map from monocular image can be categorized into supervised~\cite{fu2018deep, lee2019big}, Semi-supervised ~\cite{kuznietsov2017semi} and self-supervised ~\cite{zhou2018unsupervised, guizilini20203d, wang2018learning}. 
Self-supervised methods rely on~\cite{zhou2018unsupervised, guizilini20203d, wang2018learning}, to train the depth prediction network using Image consistency loss computed between frames of image sequence~\cite{zhou2017unsupervised}, weakly-supervised or semi-supervised methods rely on the stereopsis~\cite{kuznietsov2017semi, godard2017unsupervised}.

On the other hand, supervised methods often train convolutional auto-encoders \cite{eigen2014depth, fu2018deep}, trains a two-network stack with full LiDAR supervision, with second network refining the coarse depth predictions of the first. 
Sophisticated loss functions and backbones have been explored, such as ordinal regression loss in DORN~\cite{fu2018deep}, surface normals to model geometry \cite{li2015depth, lee2019big} and more recently transformer based~\cite{bhat2021adabins, ranftl2021vision}.  
The proposed method combines object detection~\cite{yin2020center} and supervised depth prediction~\cite{fu2018deep} into a unified single-network learning task by adding a depth prediction head to the network backbone. Results on Table~\ref{table:1} adds strength to our argument that forcing object tracking network to learn depth additionally, results in improved 2D object detection and tracking performance.

\begin{figure*}[hbt!]
 \centering
 \includegraphics[width=1.0\textwidth, keepaspectratio]{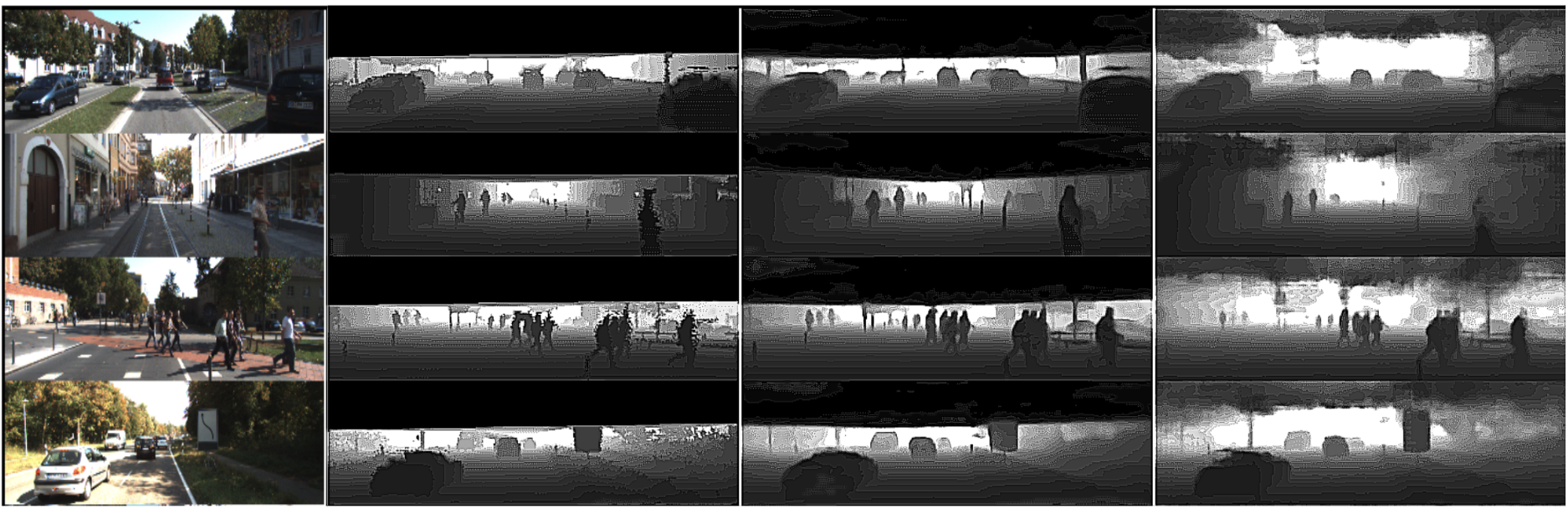}
\vspace{-0.2in}
 \caption{Qualitative results of the proposed architecture on KITTI dataset~\cite{geiger2013vision}. From left to right: input image, interpolated ground truth depth maps, predicted depth maps using proposed method, and depth predictions of DORN~\cite{fu2018deep}. }
 \vspace{-0.2in}
\end{figure*}

\noindent{\textbf{Multi-Object Tracking:}}
Several methods have attempted coupling tracking and object detection into one framework by learning them jointly~\cite{schulter2017deep, fang2018recurrent} such that the tracking algorithm can leverage both appearances and motion dynamics for modeling objects over time. Schulter {\it et al.}~\cite{schulter2017deep} models pair-wise association between objects as a function of appearance using network flow graph, Sharma {\it et al.}~\cite{sharma2018beyond} introduced reconstruction-based tracking method, and Fang {\it et al.}~\cite{fang2018recurrent} proposes a recurrent neural networks to model long and short term memory.

On the other hand, recently a few methods have been revisiting the traditional filtering based methods for data association and have shown state-of-the-art performances with them, such as ~\cite{weng2020ab3dmot, zhou2020tracking}. Weng {\it et al.}~\cite{weng2020ab3dmot} uses a simple Kalman filter \& Hungarian distance-based method to compute data association on 3D bounding boxes regressed using an object detection algorithm, and demonstrate superior performances. The proposed method extends~\cite{zhou2020tracking} architecture that performs tracking-by-detection, using a point-based framework that feeds tracklets (represented as gaussians of object centerpoints) of the previous frames back into the neural network, effectively forcing the network to learn to predict associations between two adjacent image frames.

\noindent{\textbf{Monocular BEV Object Detection:}}
Bird's Eye View (BEV)-object detection~\cite{ku2018joint} relies on global and local geometric cues of point clouds, often procured using LiDAR sensors. Although LiDAR based BEV object detection methods~\cite{ku2018joint, lang2019pointpillars, simon2018complex, yin2020center, guo2021pct} achieves higher accuracies, the cost of LiDAR sensor(s) mounted on the car during inference is exorbitant. Pseudo-LiDAR based methods~\cite{wang2019pseudo, you2019pseudo, weng2019monocular} bridge the gap between BEV and camera based methods by lifting the camera images to BEV and employ BEV object detection upon it. This involves two stages, (1) learn to regress pixel-wise depth map using techniques like~\cite{fu2018deep}, and (2) convert the depth map to Bird's Eye View and employ a BEV object detection~\cite{ku2018joint} upon it.

Unlike ~\cite{you2019pseudo, wang2019pseudo}, our method uses a single stage architecture which learns to regress both 2D Object detections and depthmap jointly, giving per pixel depth value; using regressed detection and depth maps the method renders the object as a point in 3D. While the estimated 3D object position may not necessarily be the actual center of the object, the application we are targeting does not require the accurate estimation of actual 3D object centers. Our technique overcomes the challenge of missing smaller objects, often not captured in sparse LiDAR data. We also bypass the requirement for ground-truth 3D annotations, a challenging and expensive task, by using raw LiDAR data with 2D annotations and training the network to regress depthmap (and thus {\it x,y \& z} (3D position) of the object). 

\section{Model}
\vspace{-0.05in}
\subsection{Objects as 3D Points}\label{sub:objects_as_3dpoints}
Given an input image, our goal is to perform a center-point based 2D object detection, that models each object as points in 3D and offsets in 2D, and to train the network to regress them as a function of the object's appearance. Input image is represented as $I^t$ at time frame $t$, where $I \in \mathbb{R}^{H \times W \times 3}$ with $W$ and $H$ are {\it width} and {\it height} of the input image respectively. For each image frame we have 2D object annotations where $N$ 2D bounding boxes with corners $\{x^n_1, y^n_1, x^n_2, y^n_2\}$ are represented using object centers and offsets. $C \in \mathbb{R}^2$ denotes object centers and $\bigtriangleup C \in \mathbb{R}^2$ denotes offsets of the object bounding boxes. 

Using the technique proposed by~\cite{zhou2019objects} each object annotation is converted to a heat-map $Y \in [0,1]^{\frac{W}{R} \times \frac{H}{R} \times K}$ that maintains the aspect ratio of that of the input image, but is downsized by a factor ($R$), where $K$ is the number of classes. The object heatmap is constructed by converting the center-points into gaussian kernels based on the object's bounding box size and aspect ratio. The formulation is detailed in ~\cite{zhou2019objects}. 

Given two consecutive image frames $I^t$ and $I^{t-1}$. The network regresses the heatmap $\hat{Y} \in [0,1]^{\frac{W}{R} \times \frac{H}{R} \times K}$, whose peaks represent object centers $C = \{(\hat{x}^n_1 + \hat{x}^n_2 - \hat{x}^n_1)/2, (\hat{y}^n_1 + \hat{y}^n_2 - \hat{y}^n_1)/2\}$, and offsets as $\bigtriangleup C = \{(\hat{x}^n_2 - \hat{x}^n_1)/2, (\hat{y}^n_2 - \hat{y}^n_1)/2\}$. The object detection network takes as input the image, and learns to regress object centers, offsets along with object classes and confidence scores. At inference, from the regressed heatmap $\hat{Y}$ we identify peaks, and use the object offsets and centers regressed to construct each object detections.

\subsection{Tracking Objects as 3D Points}

Each detected object $b$ is represented as $b = \{C, \bigtriangleup C, w, l, d, m \}$ where $w \in [0,1]$ and $l$ are confidence of the detection and an unique label associated with each detected object, $m$ is the object class, and $d$ is the distance between the camera center and the closest 3D object point (obtained from depth prediction network). Given a pair of images  $I^t$ \&  $I^{t-1}$, the goal is to uniquely identify each object with a label and track them across frames. For that purpose, we follow a tracking conditioned on detection based method used in~\cite{zhou2020tracking}. In order to perform tracking and detection jointly, we restructure the inputs and outputs of the network detailed under Section~\ref{sub:objects_as_3dpoints}. The network takes as input, the previous frame along with the current frame, and in addition the tracklets of the previous frame. 

Like~\cite{zhou2020tracking}, the network also takes prior object detections $\{\hat{Y}_{t-1}, \hat{Y}_{t-2}, ... \hat{Y}_0\}$, but, unlike~\cite{zhou2020tracking} where the tracklets are 2D Gaussians constructed based on the object center and bounding box sizes, in our method, we model each object entirely in 3D. We model the distributions of each object in the tracklets in 3D, and for that purpose, in addition to using the 2D centers and 2D offsets alone (as in~\cite{zhou2020tracking}), we also use the regressed depth maps to model the points in 3D. Feeding the previous frame as input along with the current allows the network to learn to model objects that appear or disappear due to change of scene or occlusion. We train the network to predict 3D displacements to compute object associations in 3D.

\subsection{Monocular Depth Prediction}
It has been shown that forcing the network to predict the global depth of the whole scene allows the network to model the scale of the objects more accurately. Depth prediction algorithms~\cite{fu2018deep} learn relative object scales more efficiently as they utilize scale context through appearance cues of global and local regions around objects to model scale. Thus the goal is to predict the whole scene's depth instead of simply regressing the object's 3D position alone. For that purpose, we add an additional convolutional head to the network that regresses a downsized depth map ${D} \in \mathbb{R}^{\frac{W}{R} \times \frac{H}{R} \times 1}$.

\subsection{Losses}
We use focal loss to train the network to learn to regress object detections $z = \{C, \bigtriangleup C, m\}$. 

\[
    L_{obj}= \frac{1}{N} \sum_{z}
\begin{cases}
    (1-\hat{Y}_z)^\alpha \cdot log(\hat{Y}_z),& \text{if } \hat{Y}_z = 1\\
    (1-\hat{Y}_z)^\beta (\hat{Y}_z)^\alpha \cdot log(1-\hat{Y}_z),              & \text{otherwise}
\end{cases}
\]

For any object regressed using 2D object detection, at time t, we find it's depth using regressed depthmap $\hat{D}$ at that location. To obtain object's displacement $\hat{d}^t =(\hat{p}^{(t-1)}_i - \hat{p}^t_i)$ where $\hat{d}^t \in \mathbb{R}^{\frac{W}{R} \times \frac{H}{R} \times 3}$, we find the difference between each object's 2D position $(xy)$ \& $depth$ at time $t$ and $t-1$, where ${p}^t_i$ \& ${p}^{(t-1)}_i$ are locations of groundtruth objects in the current and previous frames respectively. We then calculate the loss for displacement using

\[
    L_{disp}= \frac{1}{N} \sum_{i=1}^{N} | {\hat{d}^t}_{\hat{p}^t_i} - ({p}^{(t-1)}_i - {p}^t_i)|
\]

\begin{figure*}[hbt!]
 \label{fig:quantitative_results_2}
 \centering
 \includegraphics[width=1.0\textwidth, keepaspectratio]{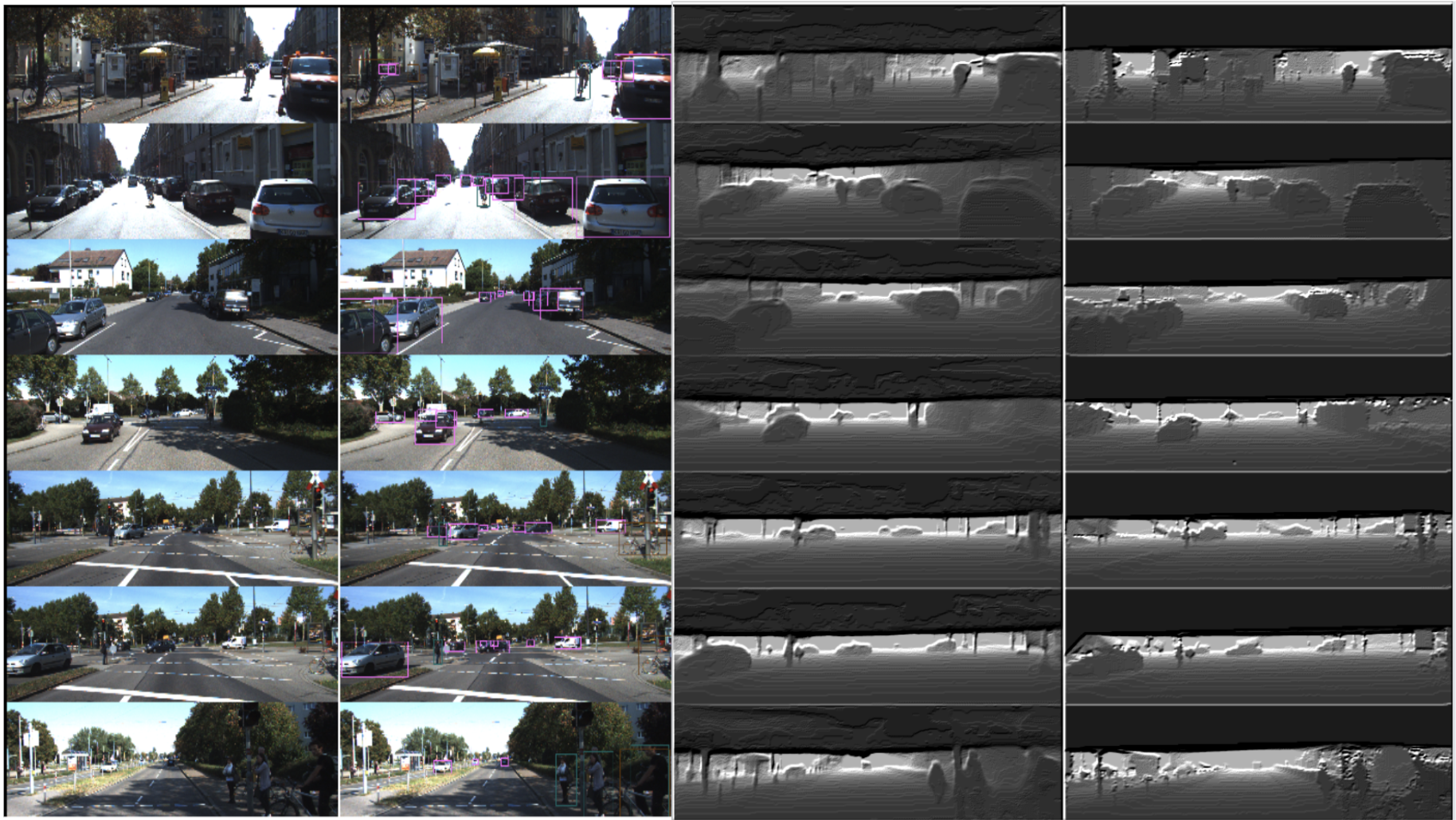}
\vspace{-0.2in}
 \caption{Qualitative results of the proposed architecture on KITTI dataset~\cite{geiger2013vision}. From left to right: input image, object detections using proposed architecture, depth predictions, and interpolated ground truth depth maps. }
 \vspace{-0.2in}
\end{figure*}

The total loss for the network training is given as:

\[
L_{total} = \alpha_1 \cdot L_{obj} + \alpha_2 \cdot L_{disp} + \alpha_3 \cdot L_{depth}
\]

where $L_{depth}$ is simply the L1 loss between regressed and ground truth LiDAR pointclouds. The lidar pointclouds are transformed \& projected into a 2D depth map ${D} \in \mathbb{R}^{\frac{W}{R} \times \frac{H}{R} \times 1}$, an input image-shaped grid (as shown in last column of Figure ~\ref{fig:quantitative_results_2}), which is then used to compute the $L_{depth} = ||D^t_{pred} - D^t_{gt}||$ where depth map $D^t_{pred}$ \& $D^t_{gt}$ are predicted and ground truth depth maps respectively.

\begin{table*}[tbhp]
\centering
\footnotesize
\begin{tabular}{c c c c c c c c c} 
 \hline
  & Time(ms)↓ / fps↑  & MOTA ↑ & MOTP ↑ & MT ↑ & ML ↓ & IDSW ↓ & FRAG ↓ \\ [0.5ex] 
 \hline\hline
 Centertrack* & 30.47/ 33 & 81.63 & 82.96 & {\bf 85.25} & 2.87 & 44 & 157 \\
 Ours (DLA34)* & 31.70/ 31.5 & 83.54 & {\bf 84.67} & 79.86 & 3.59 & {\bf 27} & {\bf 138} \\
 Ours (MobileNetV2)* & {\bf 18.20/ 55} & 84.02 & 84.09 & 76.61 & {\bf 2.51} & 99 & 208 \\
 Ours (ResNet18)* &  28.84/ 34.6 &  {\bf 84.55} & 83.32 & 76.04 &  4.6 & 75 & 189 \\

 \hline
\end{tabular}
\caption{KITTI test set evaluation results (comparison made on the class {\it Car}). * represents evaluation results of ~\cite{zhou2020tracking} reproduced by us on KITTI validation split of~\cite{zhou2020tracking}.}
\label{table:1}
\end{table*}

\subsection{Network Architecture}
The proposed network has four inputs, current and previous image frames, object tracklets and optionally the regressed depth maps of the previous frames. The encoded tensor (from the backbone) is then fed through two different network heads, an object detection head and a depth prediction head. The object detection module takes the encoded tensor (from DLA34) and feeds it through the object detection head comprises of a series of convolutional (or sparse/deformable convolutional) layers, followed by regression of bounding boxes in 2D (object centers and offsets along {\it x} and {\it y}). Similarly, the depth prediction head, feeds the tensor through a series of convolutional layers, which to the end regresses an downsized dense depth map (the size of the map is scaled down by $R = 4$ throughout this paper). To the end, the network gathers the regressed object centers and depth predictions to output 3D point (and 2D offsets), which along with the displacements is used to estimate label associations. 

\begin{table*}[tbhp]	  
\centering       
\footnotesize
\begin{tabular}{c c @{\hspace{2ex}} c @{\hspace{2ex}} c c @{\hspace{2ex}} c @{\hspace{2ex}} c @{\hspace{2ex}} c c @{\hspace{2ex}} c @{\hspace{2ex}} c}           
\toprule    
$\theta$ & \multicolumn{3}{c}{Supervision} &  \multicolumn{4}{c}{Error Metric} & \multicolumn{3}{c}{Accuracy Metric} \\
\cline{2-4} \cline{5-8} \cline{9-11}
& Depth & Pose &Unsupervised & Abs Rel & Sq Rel & RMSE & RMSE log & $\delta<1.25$ & $\delta<1.25^2$ & $\delta<1.25^3$ \\
\cline{1-11} 
DORN {~\cite{fu2018deep}} {{(50m cap)}} & \checkmark & & & 0.071 & 0.268 & 2.271 & 0.116 & 0.936 & 0.985 & 0.995\\
BTS {~\cite{lee2019big}} & \checkmark & & & 0.056 & 0.169 & 1.925 &  0.087 & 0.964 & 0.994 & 0.999\\
DPT-Hybrid\cite{ranftl2021vision} & \checkmark & & & 0.062 & - & 2.573 &  0.092 & 0.959 & 0.995 & 0.999\\
Adabins ~\cite{bhat2021adabins} & \checkmark & & & 0.058 & 0.190 & 2.360 &  0.088 & 0.964 & 0.995 & 0.999\\
AdaBins* ~\cite{bhat2021adabins} & \checkmark & & & 0.066 & 0.203 & 2.31 &  0.126 & 0.946 & 0.978 & 0.99\\
Ours (DLA34)$^*$  & \checkmark & & & { 0.102} & {0.750} & {4.137} &  { 0.169} & { 0.898} & {0.967} & { 0.986}\\
Ours (MobileNetV2)$^*$  & \checkmark & & & 0.0835 & 0.4716 & 3.6386 & 0.1508 & 0.9220 & 0.9736 & 0.9895\\
Ours (ResNet18)$^*$  & \checkmark & & & 0.1128 & 0.7137 & 4.7172 & 0.2022 &  0.8532 & 0.9427 & 0.9807\\
\cline{1-11}
PackNet-SfM ~\cite{guizilini20203d} {{(640 x 192 res.)}} & &\checkmark & & 0.078 & 0.420 & 3.485 & 0.121 & 0.931 & 0.986 & 0.996\\
Zhou {\it et al.}~\cite{zhou2017unsupervised}{({w/o exp. mask)} } &  &  & \checkmark & 0.221 &2.226& 7.527& 0.294 & 0.676 & 0.885 & 0.954\\
Zhou {\it et al.}~\cite{zhou2017unsupervised}   &  &  & \checkmark & 0.208 & 1.768 & 6.856 & 0.283 & 0.678 & 0.885 & 0.957\\  
\cline{1-11}
{\it Kuznietsov et al.}~\cite{kuznietsov2017semi} & \checkmark &   \checkmark(stereo) &  &  {0.113} & {0.741} & {4.621 }& {0.189 } &  {0.875 } & {0.964 } & {0.988 } \\
\bottomrule          
\end{tabular}  
\caption{Comparison of Monocular depth prediction results on KITTI dataset~\cite{geiger2013vision}. \textbf{*} Networks (with our implementation) trained on KITTI tracking dataset (as our method needs 2D object annotations), using only 4009 images (see Section~\ref{results_disc} for explanation), as opposed to other entries in this table that are trained using 85K images of KITTI and evaluated on our KITTI tracking split as opposed to the Eigen split~\cite{eigen2014depth} used by other methods.}    
\label{table:quantitative_results_2}   
\end{table*}

\begin{table*}[tbhp]	  
\centering       
\footnotesize
\begin{tabular}{c c @{\hspace{2ex}} c @{\hspace{2ex}} c @{\hspace{2ex}} c  c @{\hspace{2ex}} c @{\hspace{2ex}} c}              
\toprule    
$\theta$ &  \multicolumn{4}{c}{Error Metric} & \multicolumn{3}{c}{Accuracy Metric} \\
\cline{2-5} \cline{5-8}
& Abs Rel & Sq Rel & RMSE & RMSE log & $\delta<1.25$ & $\delta<1.25^2$ & $\delta<1.25^3$\\
\cline{1-8} 
Whole Image   & 0.1022 & 0.7502 & 4.1362 & 0.1685  & 0.8974 & 0.9669 & 0.9860\\ 
Object Bounding boxes  &  0.1462 & 1.4146 & 5.8425 & 0.2212 & 0.8302 & 0.9394 & 0.9705\\         
Objects within range (0m - 20m) &   0.1524 & 1.2259 & 3.2629 & 0.2157 & 0.8520 & 0.9403 & 0.9675\\ 
Objects within range (20m - 50m) & 0.1327 & 1.4691 & 6.3861 & 0.2107 & 0.8260 & 0.9386 & 0.9730\\
Objects within range  (50m - 80m) & 0.1366 & 2.4841 & 11.2572 & 0.2356 & 0.8048 & 0.9276 & 0.9592\\

\bottomrule          
\end{tabular}  
\caption{Ablation studies on depth prediction accuracy: Object regions vs Background, and depth prediction accuracy wrt. varying distances of objects from ego vehicle.}    
\label{table:quantitative_results_3}   
\end{table*} 

\section{Experiments}
\vspace{-0.05in}
We evaluate the proposed architecture on KITTI tracking dataset~\cite{geiger2012we}. To compare the performance of the proposed method against object detection, tracking and depth prediction, we evaluate our method using a series of metrics. We show the MOTA, MOTP to demonstrate the object tracking performance, in addition to using standard depth prediction metrics such as absolute and squared errors, RMSE etc, for evaluating the depth prediction performance of the proposed method.

\begin{figure*}[hbt!]
 \centering
 \includegraphics[width=1.0\textwidth, keepaspectratio]{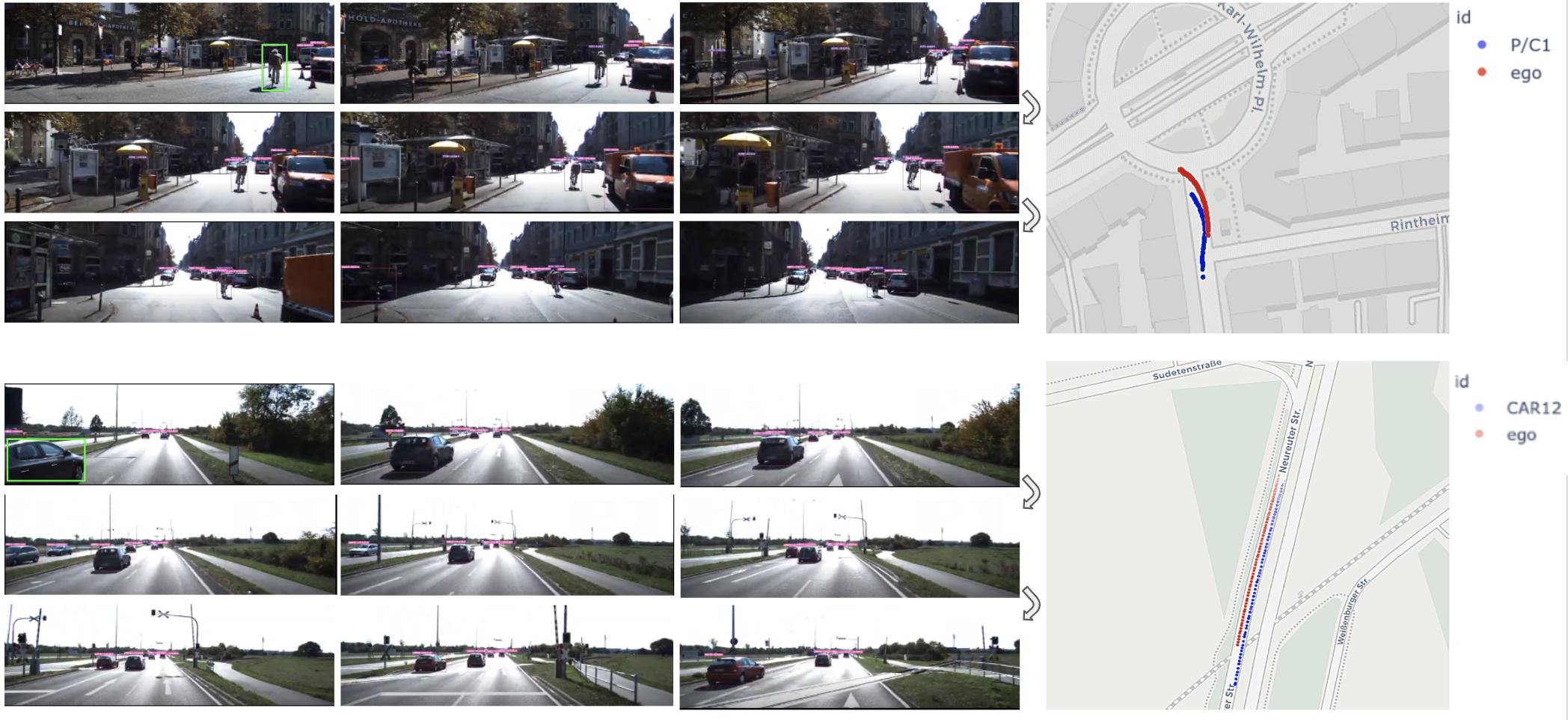}
\vspace{-0.2in}
 \caption{Qualitative representation interactions between ego vehicle and an actor of interest (marked in green): Sequence of images with its 2D object detections (bounding boxes) (left), and object 3D positions plotted on OpenStreetMaps~\cite{osm2004}. The red colored object track represents the positions of the ego vehicle, where the other actor's track is plotted in blue; each dot represents the objects' position at every video frame and the color of the dot fades over time. The ego vehicle's location is plotted using the available GPS coordinates where the actor's position is calculated relative to ego's position using the 2D object detections and depth predictions.}
 \label{fig:qual_scenario}  
 \vspace{-0.2in}
\end{figure*}


\subsection{Implementation Details}
For the backbone, we have used three different architectures - DLA34~\cite{yu2018deep}, MobileNetV2~\cite{sandler2018mobilenetv2} and Resnet 18~\cite{he2016deep}. For network training we start with the initial learning rate of $1.25\mathrm{e}{-4}$ with Adam optimizer, with exponential decay. We trained our network with a batch size 64, on Titan RTX (2080Ti) GPU and Intel Silver 4214 CPU, with stop on convergence. All the experiments reported in this paper are trained under this training setup for fair comparison. We perform several augmentations to the training data including randomly resizing images, horizontal flipping, applying affine transformations, color jittering and random cropping. For depth predictions, we projected the velodyne point cloud into a $R$-times downscaled 2D depth map. For all the experiments in this paper, we use $R=4$. While we downscale the target depth maps, we feed the input image to the network at its original resolution of 1280 x 384. 

\subsection{Dataset} 
KITTI object tracking benchmark provides 21 training and 29 test sequences. The dataset provides 2D bounding box annotations for cars, pedestrians, and 6 other classes, but only the classes {\it Car} and {\it Pedestrian} classes have been used for object detection  \cite{geiger2012we}. The KITTI tracking dataset consists of video sequences captured at 10 FPS, providing in total 8008 image frames. We split the training set into two separate subsets, with 50\% images for training and another 50\% for evaluation.

\subsubsection{Evaluation metrics}
Although KITTI benchmark evaluates tracking with the HOTA metrics, we have tried to keep the metrics similar to those reported in previous studies for a fair comparison. The common metrics used by \cite{zhou2020tracking, fu2018deep, lee2019big} are multi-object tracking accuracy, $MOTA = 1 - \frac{\sum_{t} (FP_{t} + FN_{t} + IDSW_{t})}{\sum_{t} GT_{t}}$, where $GT_{t}$ represents ground-truth bounding boxes, $FP_{t}$ represents false positives, $FN_{t}$ is false negatives and $IDSW_{t}$ identity switches; multi-object tracking precision, $MOTP = \frac{\sum_{t,i}{d_{t,i}}}{\sum_{t} c_{t}}$, where $c_{t}$ denotes number of matches in frame t and $d_{t,i}$ is the bounding box overlap of target $i$ and its assigned ground truth ~\cite{milan2016mot16}. Objects which get correctly identified by the detector but missed by tracking algorithm are considered identity switch (IDSW). Only the detections which have intersection-over-union with ground-truth $\geq$ 0.5 are taken as true positives for 2D detections. Whereas, for {\it z} (depth) tracking, bounding box center distance $\leq$ 2m on the ground plane is taken as true positive.  
We further use common error and accuracy metrics to compare our work with the previous competing works. We use Absolute relative error $ = \frac{\frac{1}{|T|} \sum_{\Bar{t}\in T}{|\Bar{t} - t|}}{t}$, Sq. Rel $ = \frac{\frac{1}{T}\sum_{\Bar{t}\in T}{||\Bar{t} - t||^2}}{t}$, RMSE $= \sqrt{\frac{1}{T}\sum_{\Bar{t} \in T}{||\bar{t} - t||^2}}$ and the log of RMSE, where T represents total available ground truth. 


\section{Results and Discussion}\label{results_disc}
\vspace{-0.05in}
Table~\ref{table:1} shows the object tracking performance on KITTI tracking dataset. In the validation set, we marginally outperform~\cite{zhou2020tracking}, one of the recent state-of-the-art methods for object tracking. The results in Table~\ref{table:1} highlight that using depth along with 2D bounding boxes improves tracking accuracy, resulting in considerably lower ID switches between objects, and also improved MOTA \& MOTP in comparison to the baseline. While both~\cite{zhou2020tracking} and the proposed method have similar computational times with DLA34, using MobileNetV2~\cite{sandler2018mobilenetv2} the computational times are significantly lower, and our method additionally regresses a high-resolution dense depth map of the whole scene. 

The depth prediction accuracy of the method is tabulated in Table~\ref{table:quantitative_results_2}. The first 4 columns represents the error (where the lower the better) and the last 3 columns show the accuracy (vice versa). 

Our models have been trained on KITTI tracking dataset (with 50:50 train/val split) out of which 4009 images are used for training and reported evaluation on the other half (since our network requires both LiDAR point clouds and 2D object annotations for training) as opposed to other methods which are trained on full KITTI dataset with over 85K images and tested on Eigen split~\cite{eigen2014depth}  
However, for fair comparison, we have also reported our experiments using AdaBins~\cite{bhat2021adabins} trained and evaluated on the same kitti tracking split that we use for our experiments. Although our performance falls short of the state-of-the-art, it has to be noted that these SOTA architectures are not suitable for real time applications, as our networks runs at between 32 - 54 fps, one of the SOTA (AdaBins) runs at 2.5 fps, far slower than our networks, on the same (2080Ti) hardware.


Table~\ref{table:quantitative_results_3} (rows 1-2) shows that the depth prediction accuracy drops for object regions (object bounding boxes, irrespective of object classes) in comparison to depth prediction estimated for the whole image (background + object bounding boxes). This is contrary to the prevalent assumption that objects (like pedestrians and vehicles) are easier to learn and model than backgrounds and thus object scale (or predicted distances) are expected to be just as if not more accurate than background. But our experiments point to contrary, proving that learning object scale is more challenging to model than most literature accounts for. The scale ambiguity is also evident from our experiments on depth prediction accuracy wrt. varying distances (rows 3-5 of Table~\ref{table:quantitative_results_3} where the squared errors are high for objects that are far away ($\geq$ 20m), the absolute errors are low, implying that proportionately more pixels are inaccurate for nearby objects whereas the errors appear amplified when objects are far away. 

Figure~\ref{fig:qual_scenario} shows the interactions between ego vehicle and an actor. The ego vehicle's position is available as GPS coordinates, where the actor's position is computed relative to the ego using the regressed depth and object detection. One of the weaknesses that we identified with our proposed method is its lack of ability to recover when encountered occlusion. Our method relies on the depth prediction to obtain the {\it z} of the object, which often corresponds to depth value at the object's 2D center. While we have an outlier removal algorithm that handles minor occlusions, the 3D estimates of objects can be affected if the object of interest is severely occluded. Such instances have been observed to have a fluctuating pattern in the {\it z}-estimates.

\section{Conclusion}
In this paper we introduced a single-stage weakly-supervised architecture that learns to detect, track and model objects in 3D using only 2D annotations. In addition to learning to detect object in 2D, our method learns to predict a more accurate scale ({\it z}) of the object in 3D. We show improved performance over~\cite{zhou2020tracking} on tracking metrics, validating our hypothesis that forcing the network to learn global scene-level scale aids 3D object detection. In addition the proposed method fairs comparably against~\cite{fu2018deep} for depth prediction. Additionally, in terms of computation, the proposed networks that run at 18{\it ms} is significantly quicker than combined run times of ~\cite{zhou2020tracking} \& ~\cite{fu2018deep} at over 400ms (30ms \& 400ms respectively). We plan to address the proposed method's shortcoming due to occlusion by adding a object-saliency~\cite{selvaraju2017grad} estimation module to our network, to robustly identify non-occluded regions of the objects and improve the overall {\it z} estimation accuracy. Also, as a natural extension to this work, we plan to remove the need for LiDAR for training data, and use self-supervised depth prediction methods while the scale of the scene can be obtained from the vehichle's IMU data, making the application completely LiDAR-free.


{\small
\bibliographystyle{ieee_fullname}
\bibliography{egbib}
}

\end{document}